\documentclass{article}
\usepackage{graphicx} 
\usepackage{booktabs}
\usepackage{multicol, multirow}
\usepackage{float}
\usepackage{verbatim}
\usepackage{cite}
\usepackage{authblk}
\usepackage{amsmath,amsfonts}
\usepackage{booktabs}
\usepackage{stfloats}
\usepackage{url}
\usepackage{rotating}
\usepackage{geometry}
\usepackage{booktabs}
\usepackage{multicol, multirow}
\usepackage{url}

\title{MASK-CNN-Transformer \\For Real-Time Multi-Label Weather Recognition}

\author[a,b]{Shengchao Chen\footnote{Work done during an internship at Guangdong-Hongkong-Macao Greater Bay Area Weather Research Center for Monitoring Warning and Forecasting (Shenzhen Institute of Meteorological Innovation). Emali: pavelchen@ieee.org.}}
\author[a]{Ting Shu\thanks{Corresponding author: shuting@gbamwf.com}}
\author[c]{Huan Zhao}
\author[d]{Yuan Yan Tang}

\affil[a]{Guangdong-Hongkong-Macao Greater Bay Area Weather Research Center for Monitoring Warning and Forecasting (Shenzhen Institute of Meteorological Innovation), Shenzhen, China}
\affil[b]{Australian Artificial Intelligence Institute, School of Computer Science, FEIT, University of Technology Sydney, Sydney, NSW, Australia}
\affil[c]{The Chinese University of Hong Kong, Shenzhen, China}
\affil[d]{Zhuhai UM Science and Technology Research Institute, Faculty of Science \& Technology, University of Macau, Macau, China}

\date{The manuscript have been accepted by Knowledge-Based Systems\footnote{https://doi.org/10.1016/j.knosys.2023.110881}}

\begin{document}

\maketitle

\begin{abstract}
Weather recognition is an essential support for many practical life applications, including traffic safety, environment, and meteorology. However, many existing related works cannot comprehensively describe weather conditions due to their complex co-occurrence dependencies. This paper proposes a novel multi-label weather recognition model considering these dependencies. The proposed model called MASK-Convolutional Neural Network-Transformer (MASK-CT) is based on the Transformer, the convolutional process, and the MASK mechanism. The model employs multiple convolutional layers to extract features from weather images and a Transformer encoder to calculate the probability of each weather condition based on the extracted features. To improve the generalization ability of MASK-CT, a MASK mechanism is used during the training phase. The effect of the MASK mechanism is explored and discussed. The Mask mechanism randomly withholds some information from one-pair training instances (one image and its corresponding label). There are two types of MASK methods. Specifically, MASK-I is designed and deployed on the image before feeding it into the weather feature extractor and MASK-II is applied to the image label. The Transformer encoder is then utilized on the randomly masked image features and labels. The experimental results from various real-world weather recognition datasets demonstrate that the proposed MASK-CT model outperforms state-of-the-art methods. Furthermore, the high-speed dynamic real-time weather recognition capability of the MASK-CT is evaluated. 
\end{abstract}

\section{Introduction}
pWeather, encompassing wind, temperature, humidity, precipitation, and other atmospheric conditions, greatly influences people's lives and societal progress~\cite{chen2023prompt,chen2023spatial}. In autonomous driving, precise weather analysis plays a vital role in making appropriate and safe decisions. By recognizing weather conditions, autonomous vehicles can adjust their driving behavior, identify and avoid hazards, optimize routes, and improve sensor fusion and perception. This integration allows them to navigate effectively under various weather conditions, ensuring passenger safety and enhancing overall driving efficiency~\cite{kurihata2006raindrop,pavlic2013classification}. Therefore, continuous monitoring of real-time weather conditions is a subject of scientific importance with significant social impact~\cite{lu2014two,li2017multi,elhoseiny2015weather,katsura2005view,chen2022dynamic}.

Image-based weather conditions analysis offers significant advantages in terms of cost and efficiency. However, most existing studies on weather condition recognition were based on hand-crafted weather features, limiting their effectiveness to specific weather conditions. Furthermore, the results are unsatisfactory when weather features are not prominent~\cite{zhao2018cnn}. Additionally, weather conditions that rely on sensor networks carrying cameras suffer from high maintenance costs and inefficient identification, making them progressively less competitive~\cite{kurihata2005rainy, roser2008classification,yan2009weather,kurihata2006raindrop,pavlic2013classification, song2014weather,li2014method,lu2014two,li2017multi}. Therefore, weather recognition through image analysis can significantly improve the efficiency and cost-effectiveness of identifying weather conditions in various applications. However, further research is needed to overcome the limitations of existing methods and to develop more reliable and accurate models.

The progress in Deep Learning (DL) and the extensive deployment of cameras has facilitated highly accurate and cost-effective recognition of weather conditions. DL-based weather recognition methods provide substantial advantages over traditional, hand-crafted feature-based approaches~\cite{chen2023tempee}. However, accurately recognizing weather conditions remains challenging due to several inherent obstacles. Firstly, the intricate and interconnected nature of weather conditions poses challenges for accurately recognizing a single outdoor image. Secondly, the available datasets for weather recognition images often exhibit undesired variations, further complicating the recognition process. Lastly, the lack of detailed information in these datasets can hinder the accuracy of weather recognition. These challenges are detailed as follows.

\begin{figure}[tbh]
    \centering
    \includegraphics[width=0.49\textwidth]{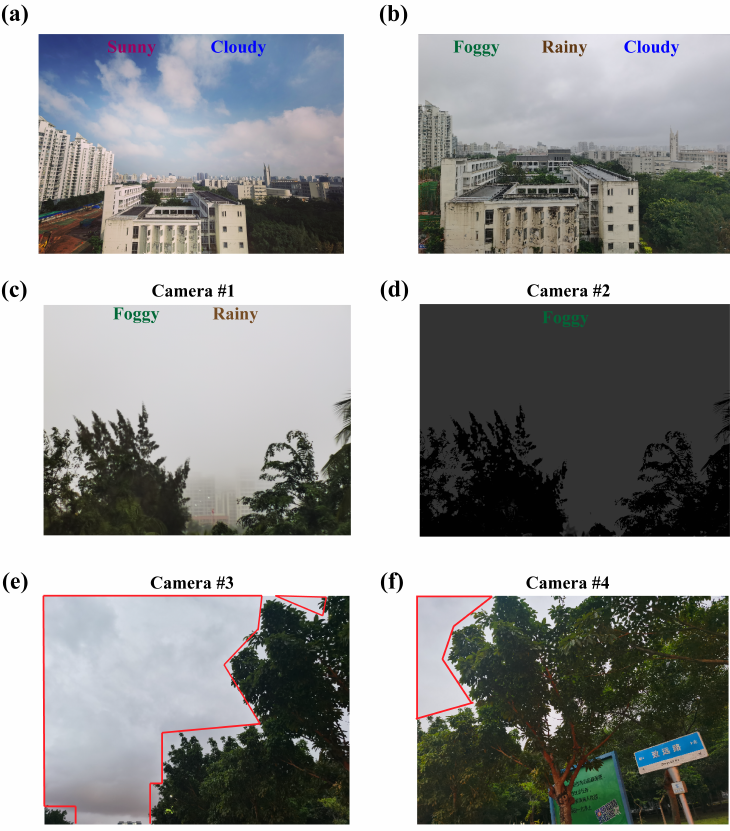}
    \caption{(a) A foggy and rainy sky taken by Camera \#1. (b) Sky taken by Camera \#2 that with different parameters from Camera \#1, the scene is the same as (a). (c) A sunny image with clouds in the sky. (d) A sky that is foggy, rainy, and cloudy at the same time. (e) Outdoor map taken by camera three, the red box represents the primary concentration area of features. (f) The outdoor image was taken by Camera \#4, and the shooting angle is not consistent with Camera \#3; the red box represents the primary concentration area of features.}
    \label{F1}
\end{figure}

Image-based weather recognition is mainly limited by a single image's complex and interwoven weather conditions. Fig.~\ref{F1}(a) and (b) show images of the same scene taken at different times, which contain at least two weather conditions. However, many early studies recognized only one type of weather from the images, so much so that they were quite limited. The attributes of the outdoor images significantly influence image-based weather recognition (e.g., light intensity, contrast, viewing angle, etc.), and differences in these attributes are generally caused by differences between shooting devices, locations, parameters, etc. 
Weather recognition based on images is predominantly constrained by the complexity and intertwining of weather conditions within a single image. Fig.~\ref{F1}(a) and (b) display images of the identical scene captured at different times, each depicting at least two distinct weather conditions. Nevertheless, numerous early studies merely recognized a single type of weather from the images, resulting in significant limitations. Various attributes of outdoor images notably influence image-based weather recognition, including light intensity, contrast, viewing angle, and others. These attribute variations often stem from disparities in shooting devices, locations, parameters, and other factors. Fig. \ref{F1}(c) and (d) are photographs of the same scene taken by the same camera at different exposure levels. Their recognition results differ because the prominent rainfall features are ignored. After all, the light intensity in Fig. \ref{F1}(b) is lower than that in Fig. \ref{F1}(a). In addition, the reduction in the area of identifiable feature area in the image due to the difference in the shooting viewpoint also affects the recognition accuracy. Fig. \ref{F1}(e) and (f) show images taken from different viewing angles in the same scene of the device. Their feature areas are mainly concentrated in the sky, while Fig. \ref{F1}(f) has a much smaller recognition area than Fig. \ref{F1}(e) due to the different shooting perspectives. Finally, the overly ideal situation of the dataset used to train a deep learning-based weather recognition model can pose a serious challenge to the usefulness of the model. Firstly, utilizing more ideal images proves beneficial for capturing significant feature regions. Secondly, the dataset could contain labeling errors, even with multiple people involved in simultaneous processing. Moreover, the current datasets primarily consist of discrete scene images. However, this contradicts real-life scenarios that are continuous, dynamic, and non-ideal. These aforementioned challenges significantly impede current weather recognition based on DL and images.

Recognizing weather conditions in outdoor images poses a significant challenge due to the co-occurrence dependence among various weather phenomena and the potential presence of multiple weather conditions within a single image. Previous studies have explored the use of graphical neural networks, recurrent neural networks, and their variants to model these dependencies. However, their reliance on predefined relationships between weather conditions hampers their competitiveness. Additionally, the recognition process is influenced differently by various regions of the image. Consequently, it becomes crucial to comprehensively evaluate the impact of different image regions on weather condition recognition.

To this end, we propose a novel architecture called MASK-CNN-Trans-former (MASK-CT) that integrates a convolutional neural network (CNN) and transformers. The CNN extracts complex weather features from the images, while transformers are used to model the dependencies between weather conditions and to explore the relationship between different regions of the image and weather conditions. The MASK strategy is used during model training to enhance its generalization ability and enable it to be applied to real-world complex weather recognition problems. Furthermore, most existing image-based weather recognition research has not considered continuous, dynamically changing scenes. Therefore, we construct a real-time weather recognition dataset to evaluate the proposed model's performance in these scenarios. Overall, our proposed MASK-CT architecture offers an effective approach to address the multi-label weather recognition task by incorporating transformers to handle the dependencies between weather conditions and CNNs to extract features from images. The evaluation on the real-time weather recognition dataset demonstrates the model's effectiveness in dealing with dynamic scenes, making it a promising solution for practical applications.

In summary, there are four main contributions of this work:
\begin{itemize}
    \item To provide more accurate and reliable recognition of complex weather patterns for related real-world applications, we propose the MASK-CNN-Transformer, which combines pre-trained CNN and Transformer models to capture potential associations between features and context in outdoor images.

    \item To improve the model's generalization ability and recognition performance, we propose and employ MASK strategy during the training phase that randomly selects regions from the image and its labels. This trains the model to establish potential relationships between global-local features and interactive weather conditions. 
    \item To validate the effectiveness of the proposed MASK-CT in dynamic, continuous real-world situations, a real-time weather recognition dataset containing continuously changing video sets in three different scenarios is constructed and estimated by the proposed model in this work.
    \item Extensive experiments on real-world datasets demonstrate the proposed MASK-CT can achieve the SOTA performance in weather recognition and it can achieve dynamic real-time weather recognition rate of up to 101.3 FPS.
\end{itemize}

The remainder of this work is in the following: Section II reviews related works about weather recognition and multi-label image recognition. Section III describes the proposed approach in detail. Section IV describes the specific implementation details of the experiment and shows the results. Section V and Section IV discuss and conclude this work, respectively.

\section{Related work}
This section presents an extensive review of the existing approaches for weather condition recognition and multi-label classification tasks. The discussion is structured into three main parts: weather recognition with hand-crafted features, weather recognition with CNNs, and multi-label classification tasks. A comprehensive summary of the research endeavors on weather condition recognition is tabulated in Table 1.

\begin{table*}[htbp]
  \centering
  \caption{Partial summary of research on image-based weather condition recognition.}
  \resizebox{\textwidth}{!}{
    \begin{tabular}{cccp{8.865em}c}
    \toprule
    Research & Approach & Years & \multicolumn{1}{c}{Label} & Framework \\
    \midrule
     Kurihata \textit{et al.}\cite{kurihata2005rainy} & Template matching & 2005 & \multicolumn{1}{c}{Rainy} & PCA+Eigendrops \\
     Roset \textit{et al.}\cite{roser2008classification} & SVM based on feature vectors & 2006 & \multicolumn{1}{c}{Rainy} & SVM \\
     Yan \textit{et al.}\cite{yan2009weather} & Based on image features using Adaboost algorithm & 2009 & \multicolumn{1}{c}{Rainy, Sunny} & Real Adaboot \\
     Kurihata \textit{et al.}\cite{kurihata2006raindrop} & Feature matching & 2006 & \multicolumn{1}{c}{Rainy} & PCA+Eigendrops \\
     Pavlic \textit{et al.} \cite{pavlic2013classification} & Based spectral features using linear classifier & 2013 & \multicolumn{1}{c}{Foggy} & / \\
     Bronte \textit{et al.} \cite{bronte2009fog} & Based on visibility distance estimation & 2009 & \multicolumn{1}{c}{Foggy} & / \\
     Lu \textit{et al.} \cite{lu2014two} & Based on weather cues using collaborative learning & 2014 & \multicolumn{1}{c}{Sunny and cloudy} & / \\
     Li \textit{et al.} \cite{li2014method} &Building decision trees and SVM & 2014 & \multicolumn{1}{c}{Clear, Fog, Overcast, and Rain} & Tree, SVM \\
     Song \textit{et al.} \cite{song2014weather} & Extract image's feature & 2013 & \multicolumn{1}{c}{Sunny, Snowy, Fog, Rain} & KNN \\
     Zhang \textit{et al.} \cite{zhang2016scene} & Based on multi-category specific dictionary learning and multi-core learning & 2016 & \multicolumn{1}{c}{Sunny, Rainy, Snowy, Haze} &  /\\
     Elhoseiny \textit{et al.} \cite{elhoseiny2015weather} & Using CNN & 2015 & \multicolumn{1}{c}{Cloudy, Sunny} & CNN \\
     Shi \textit{et al.} \cite{shi2018weather} & Use Mask R-CNN & 2018 & \multicolumn{1}{c}{Sunny, Foggy, Rainy, and Snowy} & Mask R-CNN \\
     Lin \textit{et al.} \cite{lin2017rscm} & Using Region Selection and Concurrency Model & 2017 & \multicolumn{1}{c}{sunny, cloudy, rainy, snowy, haze, and thunder} & RSCM \\
     Zhao \textit{et al.} \cite{zhao2018cnn} & Combining CNN and RNN & 2018 & \multicolumn{1}{c}{Sunny, Cloudy, Rainy, Snowy, Moist, and Rainy} & CNN-RNN \\
     Li \textit{et al.} \cite{li2017multi} & Classify weather and segmented weather cue using CNN & 2017 & \multicolumn{1}{c}{Sky (blue, gray), Shadow, and Clouds (white, dark)} & CNN \\
     Yu \textit{et al.}\cite{yu2020global} & global similarity and local salience modules-based global-similarity local-salience network & 2020 &  \multicolumn{1}{c}{Fog50, Fog200, Fog500, RoadIce, RoadSnow, RoadWet and Sunny} & CNN\\
     Xiao \textit{et al.}\cite{xiao2021classification} &  VGG16 based on dilated convolution & 2021  & \multicolumn{1}{c}{hail, rainbow, snow, rain, lightning, dew, sandstorm, frost, fog/smog, rime, and glaze} & CNN \\
     Tian \textit{et al.} \cite{tian2021weather} & Classify weather via spiking neural network & 2021 & \multicolumn{1}{c}{cloudy, rainy, sunny and sunrise} & SNN\\
     Roy \textit{et al.}\cite{roy2022awdmc} & Adversarial Weather Degraded Multi-class scenes Classification Network & 2022 & \multicolumn{1}{c}{Foggy, Haze, Dust, Rain, and Poor illumination} & Pruning-CNN\\
     Garcea \textit{et al.}\cite{garcea2022self} & Self-supervised and semi-supervised learning & 2022 & \multicolumn{1}{c}{rainy} & LSTM\\
     Samo \textit{et al.} \cite{samo2023deep} & Classify weather via Vision Transformer & 2023 & \multicolumn{1}{c}{sunny, cloudy, foggy, rainy, wet, clear, snowy and icy} & Transformer \\
     Mittal \textit{et al.} \cite{mittal2023classifying} & Classify weather based on pre-trained CNN & 2023 & \multicolumn{1}{c}{cloudy, rainy, shine, sunrise} & CNN\\
    \bottomrule
    \end{tabular}}
  \label{T1}%
\end{table*}%

\subsection{Weather recognition with hand-crafted features}
Real-time automatic weather recognition is crucial for ensuring safe driving. Several research studies \cite{kurihata2005rainy, roser2008classification, yan2009weather, kurihata2006raindrop} have employed vehicle cameras to capture images and recognize weather conditions. Raindrop features were extracted from in-vehicle photos to identify rainy weather in works by Kurihara \textit{et al.} \cite{kurihata2005rainy, kurihata2006raindrop} and Yan \textit{et al.} \cite{yan2009weather}, using the template matching method. Additionally, global features such as HSV color histograms, gradient magnitude histograms, and road information have been used to distinguish between sunny and cloudy days and haze. Roser \textit{et al.} \cite{roser2008classification} extracted feature histograms from various regions of the original image and averaged them into multiple fractions to characterize outside rainfall. Pavlic \textit{et al.} \cite{pavlic2013classification} processed the power spectrum of the image using a Gabor filter to detect haze, while Brone \textit{et al.} \cite{bronte2009fog} employed the Sobel filter to detect haze based on image edges.

Lu et al.~\cite{lu2014two} employed hand-crafted local features, such as sky, reflections, and shadows, to recognize weather conditions. On the other hand, Li et al.~\cite{li2014method} combined global features with the Support Vector Machine (SVM) and decision trees to recognize weather conditions. Song et al.~\cite{song2014weather} utilized inflection point information, image noise, edge gradient energy, power spectrum slope, and contrast saturation to assess weather conditions in outdoor images synthetically. Zhang et al.~\cite{zhang2016scene} utilized global and local features to identify the weather conditions of a single image.

Although these studies have developed various hand-crafted features for weather recognition and demonstrated promising results in specific applications, they suffer from certain limitations. Specifically, these approaches have been developed for specific conditions or perspectives, resulting in limited generality.

\subsection{Weather recognition with CNNs}
CNN have exhibited exceptional performance in a range of computer vision tasks, including image classification, target detection, and semantic segmentation. Among the most notable CNN architectures are ResNet~\cite{he2016deep}, VGGNet~\cite{simonyan2014very}, and AlexNet~\cite{krizhevsky2012imagenet}. In recent years, there has been an increase in the use of CNNs for weather recognition tasks. Elhoseiny~\textit{et al.} \cite{elhoseiny2015weather} used a fine-tuned AlexNet model to recognize dual weather conditions based on the dataset presented by Lu \textit{et al.}~\cite{lu2014two}. Shi \textit{et al.}~\cite{shi2018weather} used the VGG model to extract image foreground features for four-weather classification. Lin \textit{et al.}~\cite{lin2017rscm} proposed a CNN-based weather recognition framework, RSCM, for multi-mine weather recognition. Lu \textit{et al.}~\cite{lu2014two} combined hand-crafted weather features with a CNN model for weather classification.

However, these methods only considered the weather recognition task as a simple binary-label classification problem, ignoring the correlations among different weather conditions. As discussed in \cite{lu2014two, zhao2018cnn}, weather phenomena are complex and interdependent, and different weather conditions may co-occur. Li \textit{et al.}~\cite{li2017multi} used weather cues to assist semantic segmentation, providing a framework for describing multiple weather situations. However, this approach relies on manageable cues for humans and does not address the issue of partial weather information loss.

To address these shortcomings, Zhao \textit{et al.}~\cite{zhao2018cnn} proposed a CNN-RNN architecture that recognizes various weather conditions by considering the weather recognition task as a multi-label classification problem. However, the RNN model used a predefined order for predicting weather conditions, limiting its flexibility. In addition, some studies use expansion of the convolution module~\cite{yu2020global,xiao2021classification,mittal2023classifying} or more flexible ways of normalizing networks~\cite{roy2022awdmc} to identify road weather conditions.

\subsection{Multi-label classification task}
Multi-Label Classification represents a extensively researched challenge within the domain of computer vision. To attain optimal prediction performance, models must explicitly account for label dependencies. Recognizing multiple labels in an image presents a formidable challenge; however, recent research has shown considerable advancements in this field. The existing literature on Multi-Label Classification can be categorized into four main groups: multi-classifier fusion, conditional label inference, shared embedded space, and label graphics modeling. Each of these categories will be concisely introduced in the following.
\paragraph{Multi-classifier fusion} some approaches used deep CNNs to implement multi-label classification by building a single classifier for each category and fusing the results \cite{wang2020research, al2022detection, kukreja2022weathernet}. Such methods ignored the dependency among various labels and were not competitive in the current multi-label classification tasks. 

\paragraph{Conditional label inference} Autoregressive models \cite{dembczynski2010bayes, read2011classifier, nam2017maximizing, wang2016cnn} used the chain rule to estimate the actual joint probability of the output labels given the inputs. Similar with the multi-classifier fusion, the drawback of such methods is that they predict one label at a time and need to provide a pre-assembled set of labels, whose execution efficiency was limited. 

\paragraph{Shared embedded space} This class of methods uses input features and output labels projected onto a shared latent embedding space to accomplish multi-label classification\cite{yeh2017learning, bhatia2015sparse}. However, these methods are still limited by pre-defined complex relationships.

\paragraph{Label graphics modeling} Modeling label relevance through graphs has proven to be an effective approach. Several recent studies
have used Graph Neural Networks (GNN) to model label dependence and obtained good results \cite{chen2019multi, lanchantin2019neural, chen2019learning, chen2020knowledge}. However, all those methods need pre-defined label co-occurrence statistics to form a knowledge-based graph. 

In summary, weather recognition based on hand-crafted features has limited applicability to specific weather situations. This kind of approach is constrained by the production of weather features, making it challenging to achieve acceptable performance in complex environments. In contrast, CNN-based weather recognition faces the challenge of constructing co-occurrence dependencies for multiple weather situations. To address this challenge, some multi-label classification methods have been proposed, which can partially alleviate the difficulty of building weather co-occurrence dependencies. However, these methods suffer from reduced effectiveness and efficiency.

To address these limitations, we introduce a novel method that harnesses the capabilities of both CNN and Transformer models. The proposed approach neither necessitates prior knowledge nor predefined label dependencies. It inherently learns feature-label and label-label relationships. Through the fusion of CNN and Transformer strengths, our method achieves remarkable performance in weather recognition tasks, particularly in complex environments

\section{MASK-CNN-Transformer (MASK-CT)}
\begin{figure*}[tbh]
    \centering
    \includegraphics[width=0.80\textwidth]{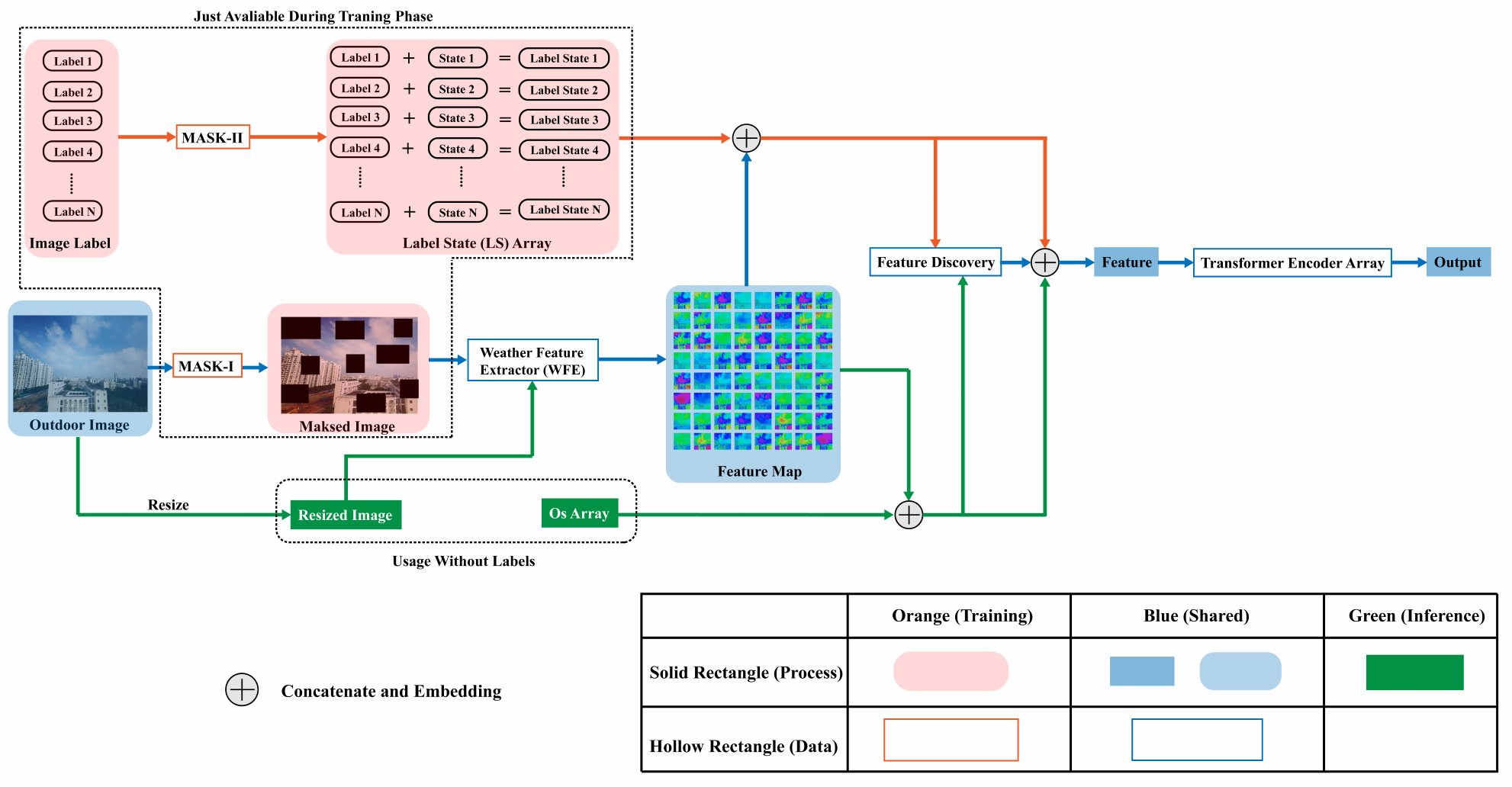}
    \caption{Illustration of the proposed MASK-CT architecture for multi-label weather condition recognition, where the label array consists of $N$ labels from the outdoor image, the black dotted box containing Mask-I and Mask-II means that these are just available during training phase. }
    \label{F2} 
\end{figure*}

The proposed architecture of MASK-CT is depicted in Figure \ref{F2}, and it consists of several components: MASK-I, a Weather Feature Extractor (WFE), MASK-II, feature discovery, and a Transformer Encoder Array. 

MASK-I is utilized for data augmentation, wherein outdoor images of different scales are divided into batches of subgraphs and are then fed into the network. These subgraphs are randomly masked with a constant-sized frame to increase the dataset's variability. The WFE extracts weather features from the images, while MASK-II randomly masks some labels corresponding to the given outdoor images with a certain probability.

Subsequently, the embeddings of the masked labels are concatenated with the feature map embeddings extracted by the WFE. The output embeddings from the feature discovery component are then fed into the Transformer Encoder Array. The Transformer Encoder Array implicitly models weather feature-weather label and weather label-weather label dependencies by means of its internal multi-headed attention module of feature associations, without providing predefined feature associations.

The final output of MASK-CT is the probabilities corresponding to the individual weather labels of the input image. In this way, the architecture comprehensively models the co-occurrence dependencies between weather conditions and the relationships between weather features. Furthermore, it also models the complex relationships between different weather features, as each feature region of a single image is linked.

\subsection{Mask-I}
\begin{figure*}[tbh]
    \centering
    \includegraphics[width=0.85\textwidth]{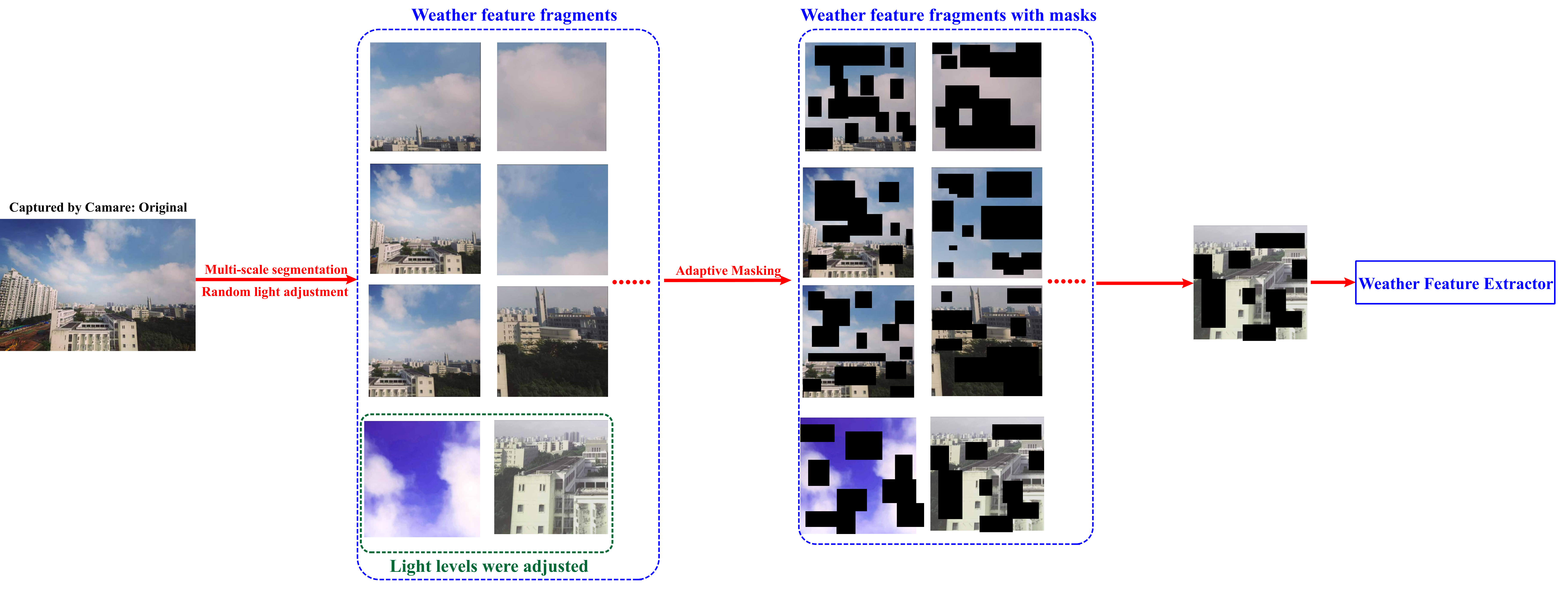}
    \caption{Implementation process of MASK-I. The original image was randomly cropped to get an image array called weather feature fragment, where the green dashed box is the feature fragment after the light level was adjusted. Note that one of the weather feature fragments with masks was randomly selected when input to the weather feature extractor of the proposed model.}
    \label{F3}
\end{figure*}

A practical and straightforward approach named MASK-I was proposed, as one of the MASK components, to improve the generalization performance and weather condition recognition accuracy of the model under different visibility, light intensity, etc. The implementation of MASK-I is shown in Fig. \ref{F3}. The image from the camera was cropped to different scales. Given the number of cropped images as $\mathcal{L}$, $0.25\mathcal{L}$ images were randomly selected to adjust the brightness (e.g., brightness, contrast, saturation, and hue), the process can be formulated as Eq.(1) to Eq.(4). First, the contrast of the image is adjusted according to the set contrast increment $\beta$:
\begin{equation}
I_{rgb}'=I_{rgb}+\frac{(I_{rgb}-T)*\beta}{I_{max}},
\end{equation}
where $I_{rgb}$ and $I_{rgb}'$ represents the $R$, $G$, and $B$ components of the image before and after being adjusted for contrast, respectively, and $T$ is the given adjustment threshold. The light of the image $L$ can be obtained according to the RGB space:

\begin{equation}
L = \frac{1}{2}*[Max(I_{rgb}')+Min(I_{rgb}')],
\end{equation}
where the $Max(I_{rgb}')$ and $Min(I_{rgb}')$ represents the maximum and minimum values of $R$, $G$, $B$ values in RGB space. The adjusted image light intensity $L'$ can be obtained from:
\begin{equation}
L' = L_{Ave.}+(L-L_{Ave.})*(\alpha+1),
\end{equation}
where $L_{Ave.}$ is the average light intensity of image, $\alpha$ is the adjustment range. The image saturation $S$ can be obtain from:

\begin{equation}
\mathrm{S}=\left\{\begin{array}{cl}
\frac{Max(I_{rgb}') -Min(I_{rgb}')}{Max(I_{rgb}') +Min(I_{rgb}')} & \text { if } L' \leq 0.5 \\
\frac{Max(I_{rgb}') -Min(I_{rgb}'}{2-(Max(I_{rgb}') +Min(I_{rgb}'))} & \text { if } L'> 0.5
\end{array}\right.
\end{equation}

\begin{equation}
    \begin{aligned}
        S' = \frac{Max(I_{rgb}') -Min(I_{rgb}')}{Max(I_{rgb}') +Min(I_{rgb}')}*L' \\
+ \frac{(1-L') * Max(I_{rgb}') -Min(I_{rgb}')}{2-[Max(I_{rgb}') +Min(I_{rgb}')]}.
    \end{aligned}
\end{equation}

As a result of the above, the image's saturation changes so that the hue of the image changes and the lighting adjustment in Mask-I was completed.

These images were then subjected to adaptive masking based on selecting features with significant differences (e.g., clouds and blue sky) for random masking in their respective regions. The principle of adaptive masking is as follows: given a square box of size $\mathcal{D} \times \mathcal{D}$, advance on the image in steps = $\mathcal{D} / 2$. First, the average intensity of the region was calculated, and the regions with more significant intensity than the average value will be masked by the black square box with the side length of $\mathcal{D} / 2$.

\subsection{Weather feature extractor (WFE)}
\begin{figure}[tbh]
    \centering
    \includegraphics[width=0.49\textwidth]{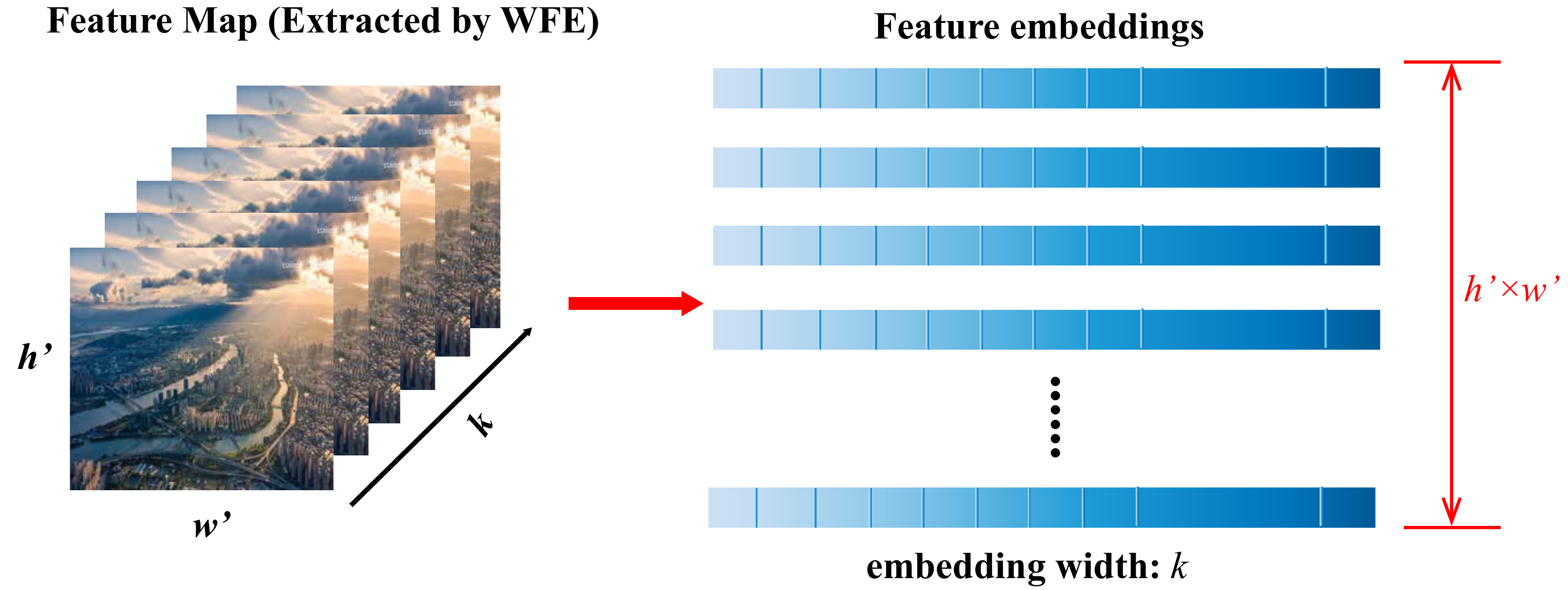}
    \caption{Shape change of weather features extracted by WFE before input to Transformer Encoder.}
    \label{F4}
\end{figure}

The initial six layers of ResNet served as the Weather Feature Extractor (WFE) responsible for extracting weather-related features from outdoor images. Considering an input image size of $h\times w \times d$ and a feature map size of $h' \times w'\times k$ after WFE, we perform feature embedding for these maps. This facilitates the Transformer encoder array in comprehending the correlation between specific weather cues and corresponding labels. The feature map embedding size is denoted as $(h' \times w', k)$, representing the subregion of the patch mapped back to the original image space. Refer to Fig. \ref{F4} for visual illustration.

\subsection{Mask-II}
\begin{figure*}[tbh]
    \centering
    \includegraphics[width=0.8\textwidth]{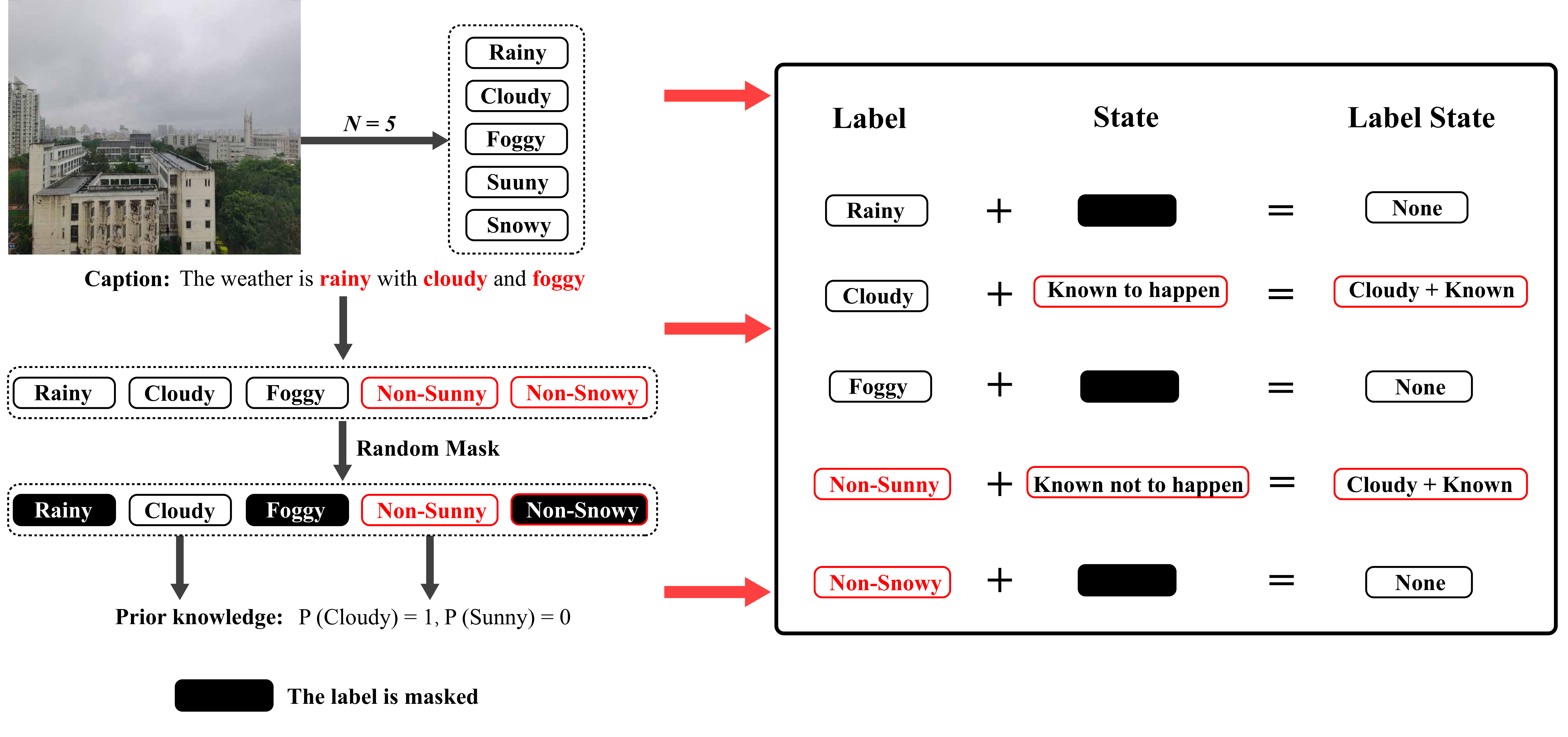}
    \caption{Illustration of the implemention of MASK-II.}
    \label{F5}
\end{figure*}

The introduction of MASK-II aims to enhance the model's flexibility in establishing inter-label connections. This approach involves masking certain labels during training, thereby encouraging the model to utilize the remaining weather labels to infer the masked ones. Figure \ref{F5} illustrates the implementation of MASK-II, wherein $N=5$, and three out of the five weather labels (rainy, foggy, and snowy) were randomly selected and marked as "Masked." Meanwhile, the remaining labels (cloudy and sunny) were regarded as "Known," with a probability of one for cloudy and zero for clear. These "known" weather labels were further categorized as "known to happen" and "known not to happen." Subsequently, the state embedding of each label was obtained by summing the embedding with its corresponding state embedding, which was then utilized as input to the transformer encoder array. The mathematical expression for Mask-II can be seen in Eq. (\ref{EQ1}).

\begin{equation}
    \mathbf{LS}_{i} = \mathbf{L}_{i} + \mathbf{S}_{i},
\label{EQ1}
\end{equation}
where $\mathbf{LS}$ denotes the label state embedding that was as input to transformer encoder array, $\mathbf{L}$ denotes the weather label, and $\mathbf{S}$ takes on the one of three possible states: Masked, Known to happen, and Known not to happen. 

\subsection{Transformer encoder array}
\begin{figure*}[tbh]
    \centering
    \includegraphics[width=0.90\textwidth]{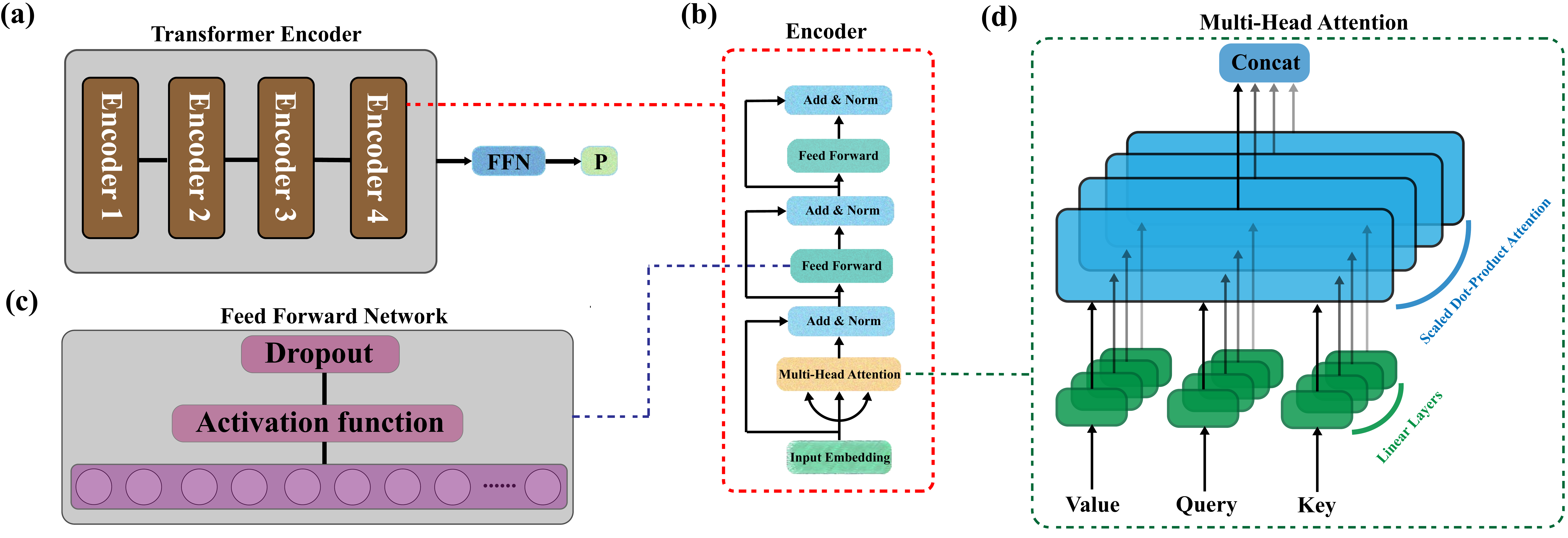}
    \caption{The architecture of Transformer Encoder. (a) Overall architecture. (b) A diagram of the Encoder's internal construction consists of a multi-headed attention module and two feed-forward networks. (c) The feed-forward network architecture in Encoder contains 2048 neural, an activation layer, and a dropout module. (d) The construction of multi-head attention in Encoder, which consists of a linear layer, Scaled Dot-Product Attention, notes that four heads were used in our work.}
    \label{F6}
\end{figure*}

The Transformer encoder array primarily comprises a multi-headed self-attention module, aimed at capturing features and their interrelationships regarding weather conditions. Additionally, it includes a forward feedback network responsible for further encoding and learning processes. The structure of the Transformer Encoder we used is shown in Fig. \ref{F6}. Given that the input Transformer Encoder embedding is $\mathbf{H}=\left\{\boldsymbol{F}_{1}, \ldots, \boldsymbol{F}_{H \times W}, \mathbf{FD}, \boldsymbol{LS}_{1}, \ldots, \boldsymbol{LS}_{i}\right\}$, where $\mathbf{FD}$ is Feature Discovery, which embeds the feature embedding extracted by CNN and the label state array into convolution operation to further explore the relationship between weather features and weather labels, which can be established as Eq.(\ref{EQ2}). In addition, where the importance of $\mathbf{H}_{i} \in \mathbf{H}$, $\mathbf{H}_{j} \in \mathbf{H}$ for each was obtained based on the multi-headed self-attention layer. The attention weights between $\mathbf{H}_{i}$, $\mathbf{H}_{j}$ named $Att_{i, j}$ can be formulated as Eq.(\ref{EQ3}). After computing the attention weights $Att_{i, j}$ for all $\mathbf{H}_{i}$, $\mathbf{H}_{j}$ pairs, we used the weighted sum to change each $\mathbf{H}_{i}$ to $\mathbf{H}_{i}'$. They can be expressed as Eq.(\ref{EQ4}) and Eq.(\ref{EQ5}).

\begin{equation}
\mathbf{FD}=Embedding(\mathbf{F}) \otimes Embedding(\mathbf{LS}),
\label{EQ2}
\end{equation}

\begin{equation}
\operatorname{Att_{i,j}}=\operatorname{softmax}\left(\frac{\mathbf{Q} \mathbf{K}^{T}}{\sqrt{d_{k}}}\right),
\label{EQ3}
\end{equation}

\begin{equation}
\overline{\boldsymbol{H}}_{i}=\sum_{i=1}^{M} Att_{i,j} \mathbf{V} \boldsymbol{h}_{j},
\label{EQ4}
\end{equation}

\begin{equation}
\boldsymbol{H}_{i}^{\prime}=\operatorname{ReLU}\left(\overline{\boldsymbol{H}}_{i} \mathbf{W}^{r}+\boldsymbol{b}_{1}\right)
\mathbf{W}^{o}+\boldsymbol{b}_{2},
\label{EQ5}
\end{equation}
where $\mathbf{Q}$ is the Query that can be expressed as $\mathbf{W}^q \cdot H_{i}$, $\mathbf{K}$ denotes the Key that can be formulated as $\mathbf{W}^k \cdot H_{j}$, $\mathbf{V}$ denotes the Value that can be formulated as $\mathbf{W}^v \cdot H_{j}$. In addition, the $\mathbf{W}^q$, $\mathbf{W}^k$, $\mathbf{W}^v$, $\mathbf{W}^r$, and $\mathbf{W}^o$ is represents the Query weight matrix, Key weight matrix, Value weight matrix, and two transformation metrics, respectively, the $\mathbf{b}_{1}$ and $\mathbf{b}_{2}$ are bias vector of $\mathbf{W}^r$, and $\mathbf{W}^o$. The proposed Transformer Encoder contains four encoder layers inside. The above update process can be repeated from layer to layer, and the updated $\mathbf{H}_{i}'$ was fed to the next encoder to continue repeating the above steps. Moreover, the obtained weights were not shared between layers.

The input $\mathbf{H}$ was transformed into $\mathbf{H}'=\left\{\boldsymbol{F}_{1}', \ldots, \boldsymbol{F}_{H \times W}', \mathbf{FD}', \boldsymbol{LS}_{1}', \ldots, \boldsymbol{LS}_{i}'\right\}$ after encoding by Transformer Encoder. Transformer Encoder has modeled the dependencies between weather features and labels and label-label dependencies. An independent feed-forward network (FFN) was used to complete the weather condition classification at the end of the Transformer Encoder. This FFN contains only a single linear layer, and its classification results $\mathbf{P}$ can be formulated into Eq.(\ref{EQ6}) and Eq.(12).

\begin{equation}
P=\mathrm{FFN}\left(\boldsymbol{LS}^{\prime}\right)=Sigmoid\left(\left(\mathbf{W}^{c} \cdot \boldsymbol{LS}^{\prime}\right)+b\right),
\label{EQ6}
\end{equation}
\begin{equation}
Sigmoid (x) =  \frac{1}{1+e^{-x}}.
\end{equation}

\section{Experiment}
This section describes the experimental setup, dataset, and the results of the conducted experiments. It involves evaluating the proposed MASK-CT on two publicly available weather recognition datasets and a self-built dataset tailored to test the model's recognition ability in dynamic scenes. Additionally, we present the results of ablation experiments on the proposed MASK-CT, where the MASK component is excluded, using the same two publicly available weather recognition datasets.

\subsection{Experimental settings}
All models and algorithms in this work are built on the Pytorch \cite{paszke2019pytorch}. ResNet152, pre-trained by ImageNet, was used as the CNN backbone in the MASK-CT to speed up the convergence of the model. Its output layer was replaced with a multi-label classification frame suitable for this task. The Transformer encoder array was trained from scratch based on the CNN backbone's output (feature map), feature discovery's output, and the label state array. During training, $\mathcal{D}$ in MASK-I was set to 18, and the MASK-II was randomized so that 25 \% labels were masked. The Transformer encoder array takes an Adam optimizer with first and second momentum of 0.9 and 0.999, respectively, to minimize the loss function. The loss function of the network can be built as Eq. (\ref{EQ7}).  
\begin{align}
        \operatorname{L}(x, y) =-\frac{1}{C} * \sum_{i} y[i] * \log \left((1+\exp (-x[i]))^{-1}\right) \\
        + (1-y[i]) * \log \left(\frac{\exp (-x[i])}{(1+\exp (-x[i]))}\right),
\label{EQ7}
\end{align}

where $x$ and $y$ denotes the ground truth and models' output, respectively. 

In addition, an exit operation with a probability of 0.35 was applied after each fully connected layer to avoid overfitting. For the integrated training of MASK-CT, the initial learning rate was set to $1e^{-5}$, and a strategy of decreasing the learning rate when the metrics stopped improving was used. Before MASK-CT was trained, each image in the dataset was resized to 384 × 384, and random noise was used for dataset augmentation. During training, each mini-batch contains 32 randomly scrambled images.

\subsection{Data}
Two available datasets used in this work are the transient attribute dataset and the multi-label weather classification dataset. A detailed description of them is given below. In addition, a self-built dataset for validating the model's ability to recognize weather in real time under dynamic scenarios is presented and described below.

\begin{figure*}[tbh]
    \centering
    \includegraphics[width=0.95\textwidth]{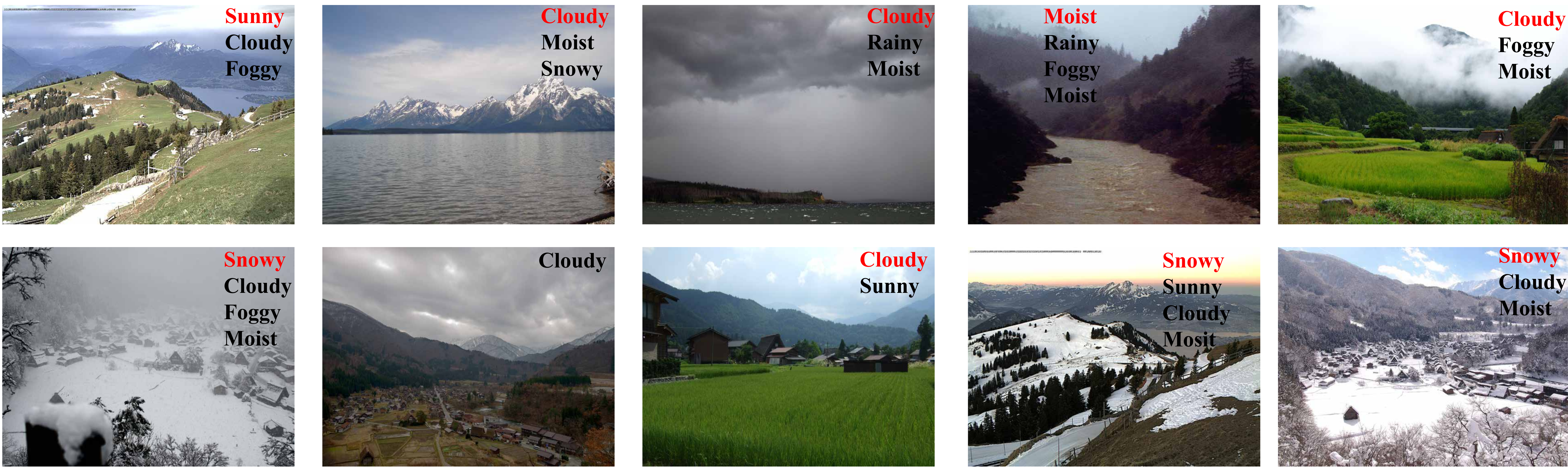}
    \caption{Example illustration of the transient attribute dataset used in this work. The weather labels belonging to each image are realistically labeled on each image, where the red label indicates the weather level with the maximum intensity.}
    \label{F7}
\end{figure*}

\begin{figure*}[tbh]
    \centering
    \includegraphics[width=0.8\textwidth]{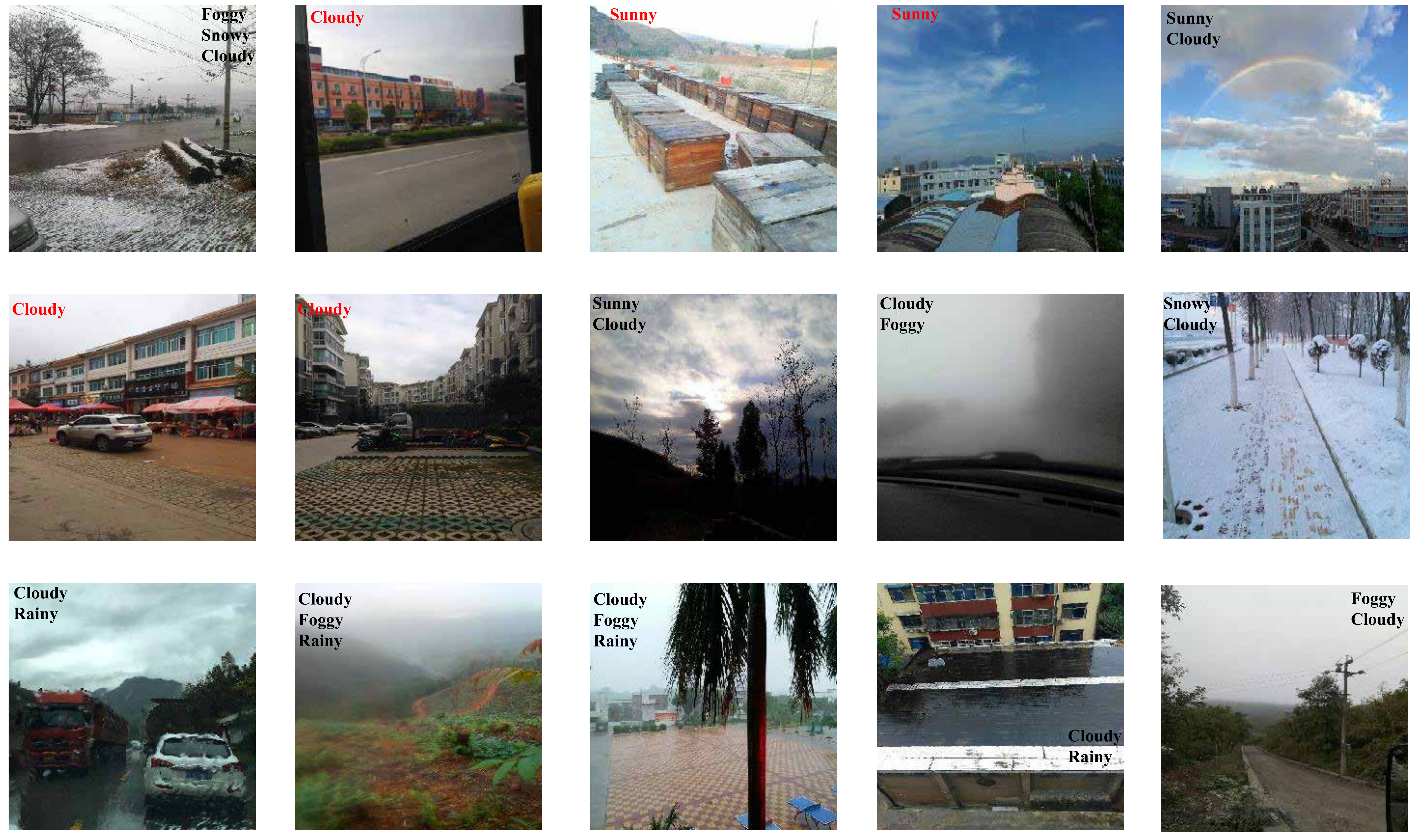}
    \caption{Example illustration of the multi-label weather dataset used in this work, where a red label means that the image has only one weather label, while a black label means that the image has at least one weather label (multiple weather conditions).}
    \label{F8}
\end{figure*}

\paragraph{Transient attribute dataset} The dataset used in this study was originally sourced from the transient attribute dataset \cite{laffont2014transient}, and was adapted for use in outdoor scene comprehension and editing. This diverse dataset comprises of images captured predominantly in outdoor settings, such as cities, towns, mountains, and lakes, with varying scales and perspectives, thus providing cross-scene diversity. To facilitate this study, the dataset was re-annotated, with all labels apart from 'sunny', 'cloudy', 'foggy', 'rainy', 'snowy', and 'moist' being removed. In the original dataset, labels were annotated with intensities, but for this study, labels with intensities greater than or equal to 0.5 were annotated as 1, while those below were annotated as 0, for ease of training. It is worth noting that some images in the dataset have a very low or even non-existent attribute intensity, particularly those captured during darker light intensities like dawn, dusk, and night. The resultant dataset contains 8571 images across 7 different weather classes. Fig. \ref{F7} illustrates an example of the dataset.
\paragraph{Multi-label weather dataset} The dataset is from Zhao~\textit{et al.} \cite{zhao2018cnn} and contains 10,000 images covering five common weather conditions in daily life such as sunny, cloudy, foggy, rainy, snowy, and including urban, suburban, and rural scenes. Each image has a different scale and perspective and has at least one weather label. An example illustration of the dataset is shown in Figure \ref{F8}.

\paragraph{Real-Time weather condition recongintion test dataset} The weather condition recognition dataset described above consists of individual photos captured from diverse locations and angles, depicting varying scenes and lighting conditions. However, the weather conditions represented in these datasets are discrete, which contrasts with the continuous and dynamic nature of real-world weather conditions. As a result, an effective weather recognition model should demonstrate proficiency in recognizing datasets with discrete weather conditions and adapt to real-world scenes, commonly referred to as dynamic scene recognition. Accordingly, a novel test dataset has been curated to assess the recognition capability of the proposed model in real-world scenarios, including its processing speed.

\begin{figure}[tbh]
    \centering
    \includegraphics[width=0.49\textwidth]{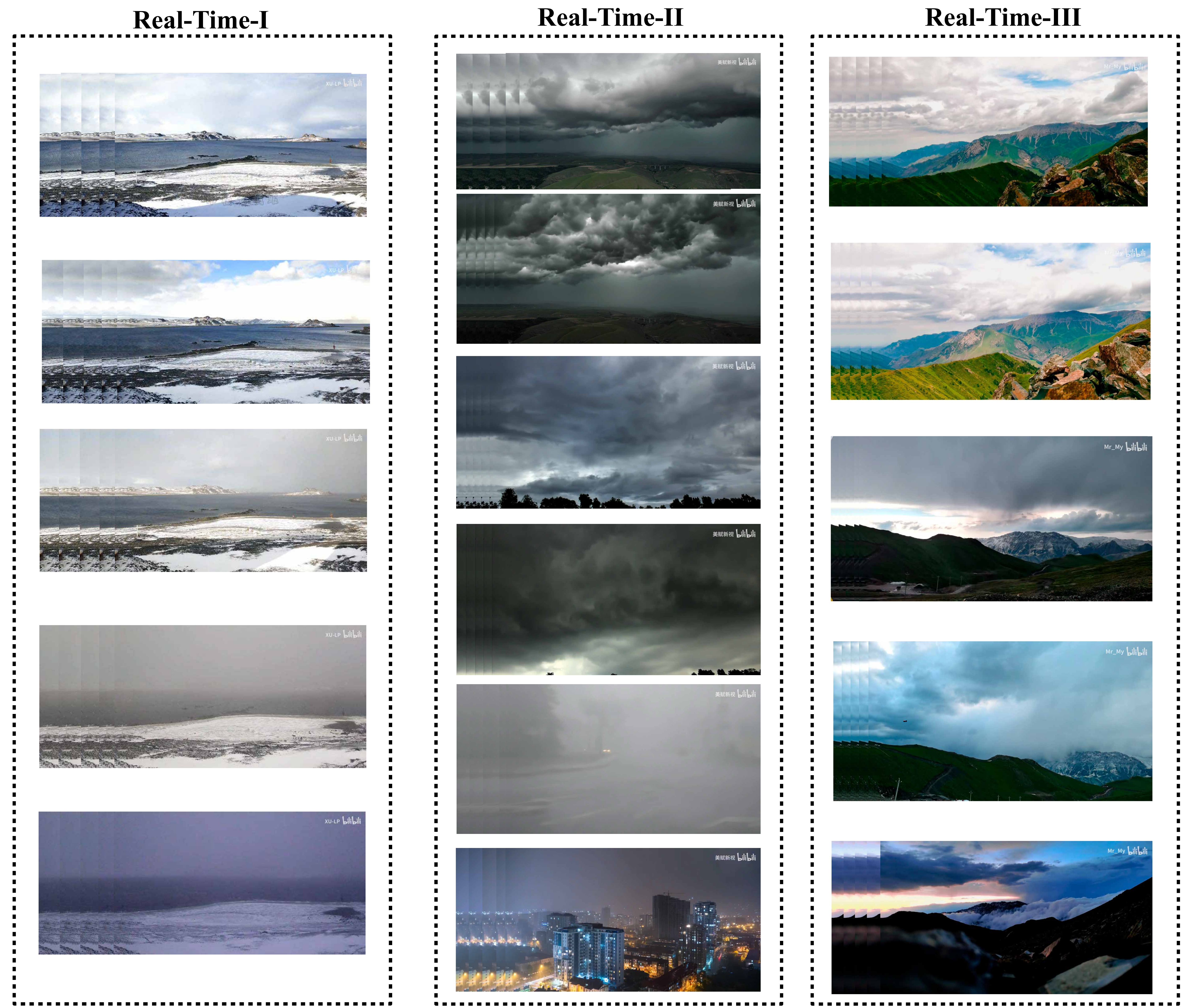}
    \caption{Example of a real-time weather condition recognition test dataset. From left to right, Real-Time-I, Real-Time-II, and Real-time-III. Real-Time-I is a video recorded at an Antarctic research station, mainly divided into five clips with different labels. Real-time-II is a video of rainstorm generation, mainly divided into six clips with different labels. Real-time-III is a video of weather changes between mountain ranges, mainly divided into five clips with different labels. They were all from bilibili that is an online video site.}
    \label{F11}
\end{figure}

The test dataset comprises of three video clips captured in the real world and is divided into three subsets, namely Real-Time-I, Real-Time-II, and Real-Time-III. These subsets were cropped frame-by-frame at a frame rate of 30 FPS. Additionally, each of these photos was comprehensively marked by a team of five markers. The elemental composition of the three subsets is illustrated in Figure \ref{F11}, and the structure of the test dataset is presented in Table \ref{T5}.

\begin{table}[htbp]
  \centering
  \caption{Composition of the real-time weather condition recognition test dataset and its details}
    \begin{tabular}{cccccccc}
    \toprule
    \multicolumn{8}{c}{\textbf{Real-time weather condition recognition dataset}} \\
    \midrule
    \midrule
    \multicolumn{2}{c}{\textbf{Subsets}} & \multicolumn{2}{c}{Real-Time-I} & \multicolumn{2}{c}{Real-Time-II} & \multicolumn{2}{c}{Real-Time-III} \\
    \multicolumn{2}{c}{\textbf{Scale}} & \multicolumn{2}{c}{2861 images} & \multicolumn{2}{c}{3426 images} & \multicolumn{2}{c}{1808 images} \\
    \multicolumn{2}{c}{\textbf{Image size}} & \multicolumn{2}{c}{640×363} & \multicolumn{2}{c}{1280×720} & \multicolumn{2}{c}{1920×1080} \\
    \bottomrule
    \end{tabular}%
  \label{T5}%
\end{table}

\subsection{Evaluation metrics and baselines}
The precision and recall of each weather label were selected as evaluation metrics. The results were first classified into the true positive (TP), true negative (TN), false positive (FP), and false negative (FN) according to the classification, as shown in Table \ref{T2}. Then, the average class precision (CP) and the average class recall (CR), which are the average of each class precision and recall, are calculated based on the above metrics, as in Eq. (14) - (17). In addition, the overall precision (OP) and overall recall (OR), which measure the actual prediction of all images in all weather classes, are also calculated, as in Eq. (18) - (20). Finally, class F1 (CF1) and overall F1 (OF1), the harmonic means of precision and recall, were formulated as Eq. (18) and Eq. (22).

\begin{table}[htbp]
  \centering
  \caption{Confusion matrix used to illustrate the TP, TN, FP, and FN.}
    \begin{tabular}{ccc}
    \toprule
    Ground Truth & \multicolumn{2}{c}{Prediction} \\
    \midrule
    \midrule
       & 1  & 0 \\
    1  & TP & FN \\
    0  & FP & TN \\
    \bottomrule
    \end{tabular}%
  \label{T2}%
\end{table}%

\begin{equation}
Recall = \frac{TP}{TP+FN},
\end{equation}

\begin{equation}
Precision=\frac{TP}{TP+FP},
\end{equation}

\begin{equation}
C R=\frac{\sum_{n=1}^{N} \sum_{n=1}^{K} Recall}{N \cdot K},
\end{equation}

\begin{equation}
C P=\frac{\sum_{n=1}^{N} \sum_{n=1}^{K} Precision}{N \cdot K},
\end{equation}

\begin{equation}
O P=\frac{\sum_{n=1}^{N} \sum_{i=1}^{K} f\left(p_{n, i}, \tilde{p}_{n, i}\right)}{N \cdot K},
\end{equation}

\begin{equation}
O R=\frac{\sum_{n=1}^{N} \sum_{i=1}^{K} f\left(p_{n, i}, \tilde{p}_{n, i}\right)}{\sum_{n=1}^{N} \sum_{i=1}^{K} p_{n, i}},
\end{equation}
where $N$ is the number of samples in the test dataset, $K$ denotes the number of weather classes, $p_{n, i}$ and $\tilde{p}_{n, i}$ denotes the actual label and predicted label of the $n^{th}$ sample on the $i^{th}$ weather class in single image, respectively. The indicator function $f(\cdot)$ is defined as Eq.(20).
\begin{equation}
f(p, \tilde{p})=\left\{\begin{array}{l}
1, \quad p=\tilde{p} \\
0, \text { otherwise }
\end{array}\right.
\end{equation}

\begin{equation}
C F 1=\frac{2\cdot CR \cdot CP}{CR+CP},
\end{equation}

\begin{equation}
O F 1=\frac{2\cdot OR \cdot OP}{OR+OP}.
\end{equation}
AlexNet~\cite{krizhevsky2012imagenet}, VGGnet~\cite{simonyan2014very}, and ResNet~\cite{he2016deep}, pre-trained by ImageNet, were chosen as benchmark models for comparison. The outputs of their classifiers were all changed to the class corresponding to the dataset (transient attribute dataset: 7; multi-label weather dataset: 5). CNN-RNN class models \cite{zhao2018cnn}, such as CNN-LSTM, CNN-ConvLSTM, and CNN-Att-ConvLSTM with an attention mechanism, were also selected. In addition, the complete graph convolutional neural network \cite{xie2021graph} with attention (GCN-A) was selected. Some popular Transformer-based image recognition like Swin Transformer~\cite{liu2021swin}, Twins Transformer~\cite{chu2021twins}, and Cross Vision Transformer (Cross ViT)~\cite{chen2021crossvit} were selected as baselines.

Considering the fact that the proposed MASK mechanism is specifically tailored for Transformer-based models due to the masked feature and real labels are embedded to Transformer Encoder, we demonstrate the effectiveness and necessity of the proposed MASK strategy in these Transformer-based models, the \textit{-CT} means that the model carry the MASK strategy. Note that these experiments are follow the same setup of MASK-CT.


\subsection{Results on the transient attributes dataset}
The transient attribute dataset was partitioned into three subsets - training, validation, and test - without any overlapping instances. The subsets were split in the proportion of 70$\%$, 10$\%$, and 20$\%$, respectively. The experiment results for the transient attribute dataset are reported in Table \ref{T3}. Based on the combined results for CP, CR, CF1, OP, OR, and OF1, the proposed MASK-CT model outperforms other models and achieves state-of-the-art performance on the dataset. Notably, the comparison between MASK-CT and CT (MASK-CT without MASK) highlights a significant gap in performance, with the latter displaying an average decrease of 5.3$\%$ compared to MASK-CT due to the lack of the MASK strategy. In addition, the performance of Transformer-based baseline with MASK strategy, such as Swin Transformer-MASK, Twins Transformer-MASK and Cross ViT-MASK are poorer relative to our proposed MASK-CT, but outperform their original form. These indicate that: (1) the effectiveness and superiority of our proposed MASK-CT model for the multi-class weather recognition in the transient attribute dataset is validated; (2) the effectiveness and necessity of the MASK strategy is further demonstrated. Furthermore, we find that recognition of rain is more challenging than other weather conditions and requires pronounced near-field features. However, the photos in the transient attribute dataset predominantly feature far-field views.

\begin{table*}[tbh]
  \centering
  \caption{Experimental result on transient attributes dataset (Per-class result: precision/recall), the \textbf{Bold} indicates the best performance.}
  \resizebox{\textwidth}{!}{
    \begin{tabular}{cccccccccccccc}
    \toprule
    Model & Sunny & Cloudy & Foggy & Snowy & Moist & Rainy & Other & CP & CR & CF1 & OP & OR & OF1 \\
    \midrule
    \midrule
    AlexNet~\cite{krizhevsky2012imagenet} & 0.756/0.892 & 0.802/0.868 & 0.688/0.688 & 0.948/0.803 & 0.840/0.903 & 0.625/0.392 & 0.789/0.224 & 0.7783 & 0.6815 & 0.7267 & 0.8967 & 0.08 & 0.8455 \\
    VGGNet~\cite{simonyan2014very} & 0.777/0.836 & 0.847/0.803 & 0.767/0.717 & 0.848/0.920 & 0.873/0.899 & 0.887/0.931 & 0.622/0.552 & 0.8022 & 0.7369 & 0.7682 & 0.9043 & 0.8155 & 0.8576 \\
    ResNet~\cite{he2016deep} & 0.805/0.832 & 0.864/0.834 & 0.756/0.727 & 0.919/0.943 & 0.936/0.897 & 0.675/0.593 & 0.675/0.593 & 0.7945 & 0.7808 & 0.7876 & 0.8519 & 0.8341 & 0.8429 \\
    CNN-LSTM~\cite{zhao2018cnn} & 0.819/0.754 & 0.883/0.555 & 0.777/0.529 & 0.654/0.205 & 0.986/0.942 & 0.271/0.373 & 0.000/0.000 & 0.6271 & 0.3653 & 0.4617 & 0.7991 & 0.3814 & 0.5163 \\
    CNN-ConvLSTM~\cite{zhao2018cnn} & 0.868/0.777 & 0.876/0.813 & 0.789/0.703 & 0.938/0.916 & 0.867/0.929 & 0.653/0.627 & 0.548/0.552 & 0.7913 & 0.7596 & 0.7751 & 0.912 & 0.8203 & 0.8637 \\
    CNN-Att-ConvLSTM~\cite{zhao2018cnn} & 0.857/0.785 & 0.851/0.852 & 0.837/0.682 & 0.952/0.896 & 0.913/0.911 & 0.656/0.454 & 0.585/0.628 & 0.8091 & 0.7428 & 0.776 & \textbf{0.9167} & 0.8231 & 0.8678 \\
    GCN-A~\cite{xie2021graph} & 0.853/0.816 & 0.859/0.858 & 0.825/0.735 & 0.94/0.908 & 0.911/0.893 & 0.763/0.651 & 0.763/0.651 & 0.8445 & 0.7754 & 0.8084 & 0.873 & 0.8342 & 0.8532 \\
    \midrule
    Swin Transformer~\cite{liu2021swin} & 0.823/0.811 & 0.732/0.729 & 0.856/0.832 & 0.912/0.894 & 0.887/0.878 & 0.742/0.723 & 0.721/0.709 & 0.8171 & 0.8494 & 0.8325 & 0.7654 & 0.8991 & 0.8280 \\
    Swin Transformer-MASK & 0.838/0.779 & 0.831/0.835 & 0.799/0.754 & 0.876/0.905 & 0.893/0.895 & 0.749/0.768 & 0.714/0.739 & 0.8194 & 0.8137 & 0.8165 & 0.8598 & 0.8706 & 0.8659 \\
    Twins Transformer~\cite{chu2021twins} & 0.825/0.770 & 0.837/0.840 & 0.762/0.745 & 0.840/0.783 & 0.885/0.875 & 0.765/0.800 & 0.735/0.755 & 0.8092 & 0.8013 & 0.8052 & 0.8487 & 0.8600 & 0.8546 \\
    Twins Transformer-MASK & 0.851/0.799 & 0.860/0.865 & 0.818/0.779 & 0.898/0.919 & 0.903/0.905 & 0.779/0.802 & 0.750/0.769 & 0.8415 & 0.8350 & 0.8382 & 0.8905 & 0.9004 & 0.8958 \\
    Cross ViT~\cite{chen2021crossvit} & 0.814/0.760 & 0.828/0.832 & 0.755/0.741 & 0.830/0.775 & 0.880/0.870 & 0.750/0.790 & 0.725/0.745 & 0.7997 & 0.7921 & 0.7959 & 0.8435 & 0.8550 & 0.8492 \\
    Cross ViT-MASK & 0.856/0.804 & 0.868/0.872 & 0.822/0.784 & 0.907/0.926 & 0.914/0.916 & 0.795/0.818 & 0.768/0.788 & 0.8377 & 0.8312 & 0.8344 & 0.8874 & 0.8975 & 0.8929 \\
    CT (MASK-CT without MASK) & 0.819/0.772 & 0.829/0.833 & 0.758/0.749 & 0.845/0.789 & 0.889/0.880 & 0.755/0.796 & 0.732/0.749 & 0.8038 & 0.7954 & 0.7996 & 0.8531 & 0.8634 & 0.8579\\
    \textbf{MASK-CT (Ours)} & 0.872/0.809 & 0.866/0.870 & 0.833/0.792 & 0.914/0.932 & 0.929/0.93 & 0.788/0.810 & 0.763/0.781 & \textbf{0.8521} & \textbf{0.8462} & \textbf{0.8491} & 0.8961 & \textbf{0.9069} & \textbf{0.9012} \\
    \bottomrule
    \end{tabular}}
  \label{T3}%
\end{table*}%

\subsection{Results on the multi-label weather classification dataset}
\begin{figure*}[h!]
    \centering
    \includegraphics[width=0.90\textwidth]{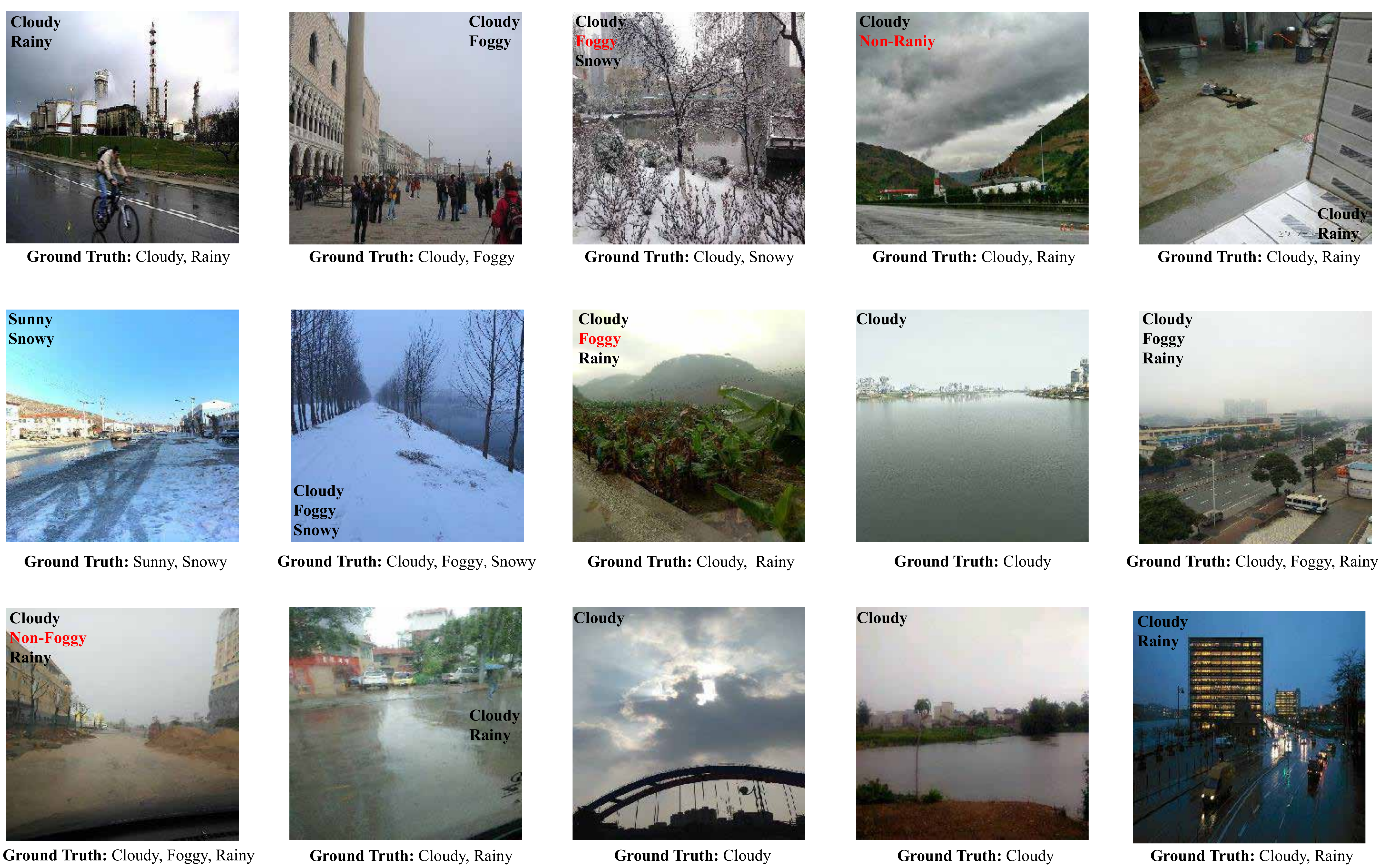}
    \caption{Weather recognition results for some images in the test dataset. Ground Truth below the image is based on the annotation of the original dataset, and the red labels in the image represent the incorrectly identified weather labels.}
    \label{F9}
\end{figure*}

The multi-label weather classification dataset~\cite{zhao2018cnn} was decomposed into training, validation, and test datasets in the ratio of 70 \%, 10 \%, and 20 \% without any intersection between the individual datasets. The some recognition results for the multi-label weather classification dataset are shown in Figure \ref{F9}. Furthermore, a comprehensive quantitative analysis for our proposed and baselines using the evaluation metrics mentioned above presented the experimental results is shown in Table \ref{T4}. Combining the results of CP, CR, CF, OP, OR, and OF1, the proposed MASK-CT showed the SOTA performance in most weather conditions, followed by Cross ViT-Mask, while Swin Transformer and Twins Transformer showed comparable performance to CT (MASK-CT without MASK), but slightly lower than our MASK-CT and Cross ViT-MASK. In addition, the proposed model degrades the performance by an average of 5.3 \% after removing MASK, which validates the effectiveness and necessity of the proposed MASK strategy and likewise shows that the proposed model has excellent performance and achieves state-of-the-art results on both public datasets. The apparent recognition performance decay between Swin Transformer/Twins Transformer/Cross ViT, Swin Transformer-MASK/Twins Transformer-MASK/Cross ViT-MASK further demonstrates the effectiveness of the MASK strategy.

The test results unveiled a remarkable phenomenon where the recognition outcomes deviated from the ground truth but aligned with the actual situation, as depicted in Figure 10. To elaborate, the ground truth was divided into two distinct categories: labels and human judgments. The former represents the labels in the original dataset, while the latter indicates the judgment output of an evaluation team comprising at least five members. Referred to as objectively misleading judgments, this issue arises from erroneous annotations in manually annotated datasets. Nevertheless, our proposed MASK-CT model proficiently identified real-world weather conditions, irrespective of the misleading labels, and made accurate predictions.

\begin{table*}[htb]
  \centering
  \caption{Experimental result on multi-label weather classification dataset, (Per-class result: precision/recall), the \textbf{Bold} indicates the best performance.}
  \resizebox{\textwidth}{!}{
    \begin{tabular}{cccccccccccc}
    \toprule
    Model & Sunny & Cloudy & Foggy & Rainy & Snowy & CP & CR & CF1 & OP & OR & OF1 \\
    \midrule
    \midrule
    AlexNet & 0.84/0.74 & 0.896/0.942 & 0.735/0.89 & 0.784/0.685 & 0.876/0.905 & 0.8263 & 0.8325 & 0.8294 & 0.9007 & 0.8668 & 0.8834 \\
    VGGNet & 0.772/0.851 & 0.927/0.915 & 0.867/0.728 & 0.814/0.701 & 0.887/0.931 & 0.8533 & 0.8252 & 0.839 & 0.9087 & 0.8494 & 0.878 \\
    ResNet & 0.903/0.719 & 0.922/0.936 & 0.841/0.855 & 0.776/0.882 & 0.947/0.938 & 0.878 & 0.8661 & 0.872 & 0.8876 & 0.8861 & 0.8868 \\
    CNN-LSTM & 0.843/0.791 & 0.897/0.958 & 0.86/0.73 & 0.83/0.694 & 0.94/0.556 & 0.8739 & 0.7458 & 0.8048 & 0.8991 & 0.8127 & 0.8537 \\
    CNN-ConvLSTM & 0.855/0.78 & 0.899/0.953 & 0.798/0.862 & 0.843/0.716 & 0.926/0.924 & 0.8643 & 0.8472 & 0.8557 & 0.9165 & 0.8793 & 0.8975 \\
    CNN-Att-ConvLSTM & 0.838/0.843 & 0.917/0.953 & 0.856/0.861 & 0.856/0.758 & 0.894/0.938 & 0.8721 & 0.8702 & 0.8705 & 0.9263 & 0.8946 & 0.9135 \\
    GCN-A & 0.881/0.851 & 0.933/0.948 & 0.902/0.840 & 0.948/0.955 & 0.925/0.796 & \textbf{0.9178} & 0.8779 & 0.8974 & 0.9222 & 0.8976 & 0.9097 \\
    \midrule
    Swin Transformer & 0.809/0.741 & 0.820/0.824 & 0.759/0.722 & 0.848/0.869 & 0.870/0.864 & 0.8032 & 0.7958 & 0.7995 & 0.8797 & 0.8903 & 0.8844 \\
    Twins Transformer & 0.823/0.766 & 0.834/0.837 & 0.769/0.752 & 0.856/0.799 & 0.899/0.889 & 0.8139 & 0.8059 & 0.8098 & 0.8533 & 0.8646 & 0.8592 \\
    Cross ViT & 0.816/0.762 & 0.830/0.834 & 0.757/0.743 & 0.834/0.779 & 0.884/0.874 & 0.8012 & 0.7936 & 0.7974 & 0.8480 & 0.8595 & 0.8537 \\
    Swin Transformer-MASK & 0.835/0.770 & 0.828/0.831 & 0.794/0.749 & 0.872/0.901 & 0.889/0.891 & 0.8163 & 0.8106 & 0.8134 & 0.8568 & 0.8676 & 0.8629 \\
    Twins Transformer-MASK & 0.848/0.786 & 0.858/0.863 & 0.814/0.775 & 0.892/0.913 & 0.897/0.899 & 0.8364 & 0.8299 & 0.8331 & 0.8868 & 0.8967 & 0.8921 \\
    Cross ViT-MASK & 0.853/0.801 & 0.865/0.869 & 0.819/0.781 & 0.902/0.921 & 0.909/0.911 & 0.8405 & 0.8340 & 0.8372 & 0.8945 & 0.9046 & 0.8999 \\
    CT (MASK-CT without MASK) & 0.872/0.864 & 0.929/0.927 & 0.866/0.840 & 0.862/0.781 & 0.910/0.877 & 0.8878 & 0.8578 & 0.8725 & 0.8849& 0.8998 & 0.8923 \\
    \textbf{MASK-CT (Ours)} & 0.902/0.909 & 0.956/0.952 & 0.878/0.890 & 0.894/0.888 & 0.940/0.950 & 0.9140 & \textbf{0.9218} & \textbf{0.9178} & \textbf{0.9419} & \textbf{0.9069} & \textbf{0.9241} \\
    \bottomrule
    \end{tabular}}
  \label{T4}%
\end{table*}%

\begin{figure*}[h!]
    \centering
    \includegraphics[width=0.90\textwidth]{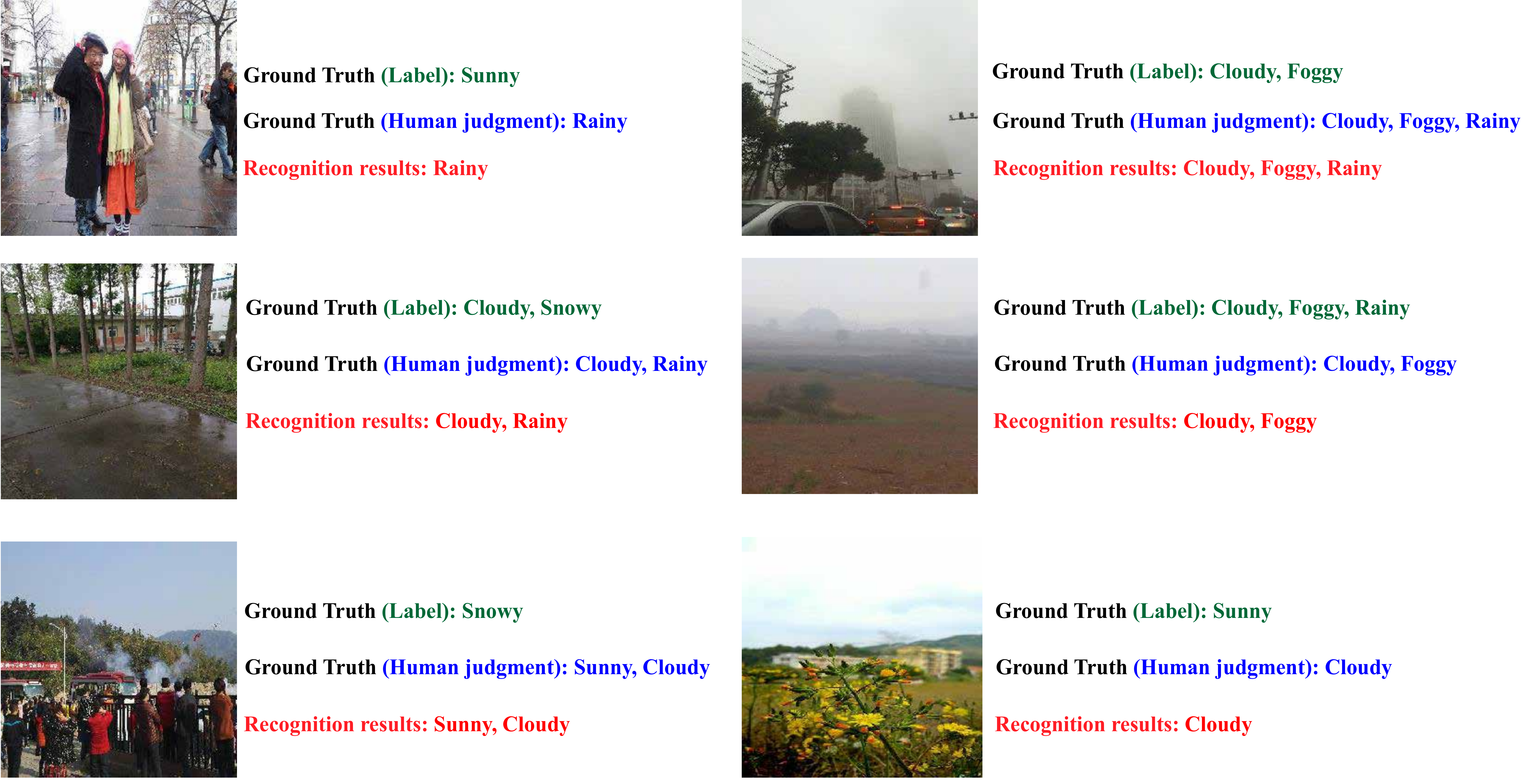}
    \caption{Some images with objectively misleading judgments, Ground Truth (Label) is the image label in the original dataset, Ground Truth (Human judgment) is the actual situation obtained by the evaluation team, and Recognition results are the proposed MASK-CT recognition results.}
    \label{F10}
\end{figure*}

\subsection{Real-time weather condition recognition}
The real-time recognition of weather conditions plays a crucial role in various applications, including transportation, agriculture, and outdoor activities. In this section, we assess the performance of the proposed MASK-CT in recognizing weather conditions in dynamic environments, aiming to evaluate its real-time capabilities and flexibility. To conduct the evaluation, we utilize a well-trained model from a multi-label weather classification dataset. We measure the model's performance on three independent subsets, namely Real-Time-I, Real-Time-II, and Real-Time-III, without any additional iterations.

To evaluate the performance comprehensively, we recorded the recognition rate while feeding the Real-Time-I, II, and III subsets into the well-trained model. The test results, as shown in Table \ref{T6}, indicate that the proposed MASK-CT achieves promising performance in all three test subsets. Furthermore, the proposed model achieves an average recognition rate of 101.3 FPS on the three subsets, which implies that it can recognize real-time weather conditions at a high recognition rate while maintaining good performance. These results demonstrate the practical value of MASK-CT for real-time weather condition recognition, and the promising performance and high recognition rate make it a valuable tool for various applications that require real-time weather condition recognition.

\begin{table*}[tbh]
  \centering
  \caption{Performance of MASK-CT trained on multi-label weather classification dataset on real-time weather condition recognition test dataset}
    \begin{tabular}{cccccccc}
    \toprule
    Test dataset & CP & CR & CF1 & OP & OR & OF1 & Frames Per Second (FPS) \\
    \midrule
    \midrule
    Real-Time-I & 0.88 & 0.802 & 0.8391 & 0.92 & 0.895 & 0.9073 & 96 \\
    Real-Time-II & 0.825 & 0.81 & 0.8174 & 0.87 & 0.84 & 0.8547 & 101 \\
    Real-Time-III & 0.891 & 0.859 & 0.8747 & 0.915 & 0.932 & 0.9234 & 107 \\
    Ave. & 0.8653 & 0.8237 & 0.8437 & 0.9017 & 0.889 & 0.8951 & 101.3 \\
    \bottomrule
    \end{tabular}%
  \label{T6}%
\end{table*}%

\section{Discussion}
Tables \ref{T3} and \ref{T4} present the performance of the proposed MASK-CT model on two different datasets: one for transient attribute classification and another for multi-label weather classification. The results of the ablation experiments demonstrate the effectiveness of the proposed MASK strategy in weather recognition. Specifically, Table \ref{T4} shows that the proposed model achieves state-of-the-art performance in recognizing weather conditions in dynamic scenes while maintaining promising results.

The enhanced performance of MASK-CT can be attributed to its capacity in effectively modeling intricate relationships among weather features, weather labels, and their interactions through the application of the Transformer Encoder. The suggested MASK strategy markedly improves the model's generalization performance by considering variations in lighting and viewpoint, which may affect recognition outcomes. Moreover, the utilization of multi-headed attention mechanisms in the Transformer Encoder allows the model to concentrate on the influence of features on labels, labels on features, and features on other features.



However, the recognition accuracy of the proposed model for images with low light intensity (e.g., darkness or dusk) may be inadequate. This is a prevalent issue across all models because the visibility of recognizable features in images captured during darkness or dusk is limited. Consequently, establishing comprehensive relationships between features and labels might be challenging for the model.

In summary, the proposed MASK-CT demonstrates promising results for weather recognition in dynamic scenes. Its strength lies in effectively modeling complex relationships between weather features and labels through the utilization of the Transformer Encoder. Nevertheless, it is important to note that the recognition accuracy might be affected by low light intensity conditions.

\section{Conclusion and Future work}
In this paper, we proposed the MASK-CNN-Transformer (MASK-CT) architecture for weather condition recognition. Our approach involved using a weather feature extractor (WFE) to extract weather features in the form of a feature map, which was then embedded along with a label state array into a Transformer encoder array. This encoder array comprised multiple encoder blocks that modeled relationships between features, labels, and feature-label pairs. To address the impact of illumination and feature regions on recognition performance, we introduced two types of MASK procedures: MASK-I and MASK-II. These procedures were used for data augmentation and partial label inference, respectively, and significantly improved the generalization performance of the model. The experimental results on transient attribute datasets and multi-label weather classification datasets demonstrated that the proposed MASK-CT model achieved state-of-the-art results. Additionally, our ablation experiment confirmed the effectiveness and necessity of the proposed MASK strategy. Furthermore, we prepared a test dataset to evaluate the real-time performance of the MASK-CT model, which achieved a detection rate of 101.3 FPS under dynamic scenarios while maintaining acceptable performance. 

Our work serves as a foundational step towards developing more comprehensive systems that incorporate weather information with other relevant factors for a holistic analysis. The experimental data in this research primarily originates from regular and surveillance cameras. In the context of autonomous driving applications, road and urban surveillance cameras can wirelessly transmit real-time images to a server, on which a pre-trained model  proposed in this paper is deployed. This model can rapidly and efficiently analyze and recognize these real-time images, and wirelessly transmit the identified weather conditions to the users, providing the current weather status for the autonomous driving vehicles. We believe that by highlighting the potential applications and future directions of outdoor weather recognition in autonomous driving, we can better emphasize the practical relevance of our work. In future work, we plan to investigate other factors that affect weather conditions and explore ways to optimize our model structure to achieve even higher accuracy and faster recognition of weather conditions.

\section*{Acknowledge}
This work was supported by the National Natural Science Foundation of China (No.42105145 and 62172458) and the Guangdong Province Natural Science Foundation (No. 2023A1515011438).

\bibliographystyle{IEEEtran}
\bibliography{mybibfile}

\end{document}